\documentclass[lettersize,journal]{IEEEtran}
\usepackage{amsmath,amsfonts}
\usepackage{algorithmic}
\usepackage{algorithm}
\usepackage{array}
\usepackage[caption=false,font=normalsize,labelfont=sf,textfont=sf]{subfig}
\usepackage{textcomp}
\usepackage{stfloats}
\usepackage{url}
\usepackage{verbatim}
\usepackage{graphicx}
\usepackage{float}
\usepackage{caption}
\usepackage{tabularx}
\usepackage{array}
\usepackage{booktabs}
\usepackage{multirow}
\usepackage{makecell}
\newcolumntype{C}{>{\centering\arraybackslash}p{3cm}}

\begin{document}

\title{High-fidelity 3D Reconstruction of Plants using Neural Radiance Field}

\author{Kewei Hu$^{1}$, Ying Wei$^{1}$, Yaoqiang Pan$^{1}$,  Hanwen Kang$^{1,\#}$, Chao Chen$^{2,\#}$  
\thanks{$^{1} $K.Hu, Y.Pan, Y.wei, and H.Kang are with College of Engineering, South China Agriculture University, Guangzhou, China}
\thanks{$^{2} $ C.Chao is with Department of Mechanical and
Aerospace Engineering, Monash University, Melbourne, Australia}}



\maketitle

\begin{abstract}
Accurate reconstruction of plant phenotypes plays a key role in optimising sustainable farming practices in the field of Precision Agriculture (PA). Currently, optical sensor-based approaches dominate the field, but the need for high-fidelity 3D reconstruction of crops and plants in unstructured agricultural environments remains challenging. Recently, a promising development has emerged in the form of Neural Radiance Field (NeRF), a novel method that utilises neural density fields. This technique has shown impressive performance in various novel vision synthesis tasks, but has remained relatively unexplored in the agricultural context. In our study, we focus on two fundamental tasks within plant phenotyping: (1) the synthesis of 2D novel-view images and (2) the 3D reconstruction of crop and plant models. We explore the world of neural radiance fields, in particular two SOTA methods: Instant-NGP, which excels in generating high-quality images with impressive training and inference speed, and Instant-NSR, which improves the reconstructed geometry by incorporating the Signed Distance Function (SDF) during training. In particular, we present a novel plant phenotype dataset comprising real plant images from production environments. This dataset is a first-of-its-kind initiative aimed at comprehensively exploring the advantages and limitations of NeRF in agricultural contexts. Our experimental results show that NeRF demonstrates commendable performance in the synthesis of novel-view images and is able to achieve reconstruction results that are competitive with Reality Capture, a leading commercial software for 3D Multi-View Stereo (MVS)-based reconstruction. However, our study also highlights certain drawbacks of NeRF, including relatively slow training speeds, performance limitations in cases of insufficient sampling, and challenges in obtaining geometry quality in complex setups. In conclusion, NeRF introduces a new paradigm in plant phenotyping, providing a powerful tool capable of generating multiple representations, such as multi-view images, point cloud and mesh, from a single process. 
\end{abstract}


\section{Introduction}
In recent years, integration of emerging sensors and Artificial Intelligence (AI) has revolutionized precision agriculture (PA), significantly enhancing the efficiency, effectiveness, and productivity of breeding and primary production in agriculture industry \cite{sishodia2020applications}. Unpredictable threats such as climate, soil characteristics, insect pests, etc. are the main challenges to maintaining and guaranteeing crop yields \cite{fu2020application}. This has given rise to the increasing importance of monitoring plant growth through the comprehensive analysis of plant phenotyping \cite{feng2021comprehensive}.
Phenomics studies a variety of phenotypic plant traits , such as growth, tolerance, yield, plant height, leaf area index, etc. \cite{asaari2019analysis}. Traditional methods for manual phenotypic measurement and analysis were labor-intensive, time-consuming, and often destructive\cite{li2020review,feng2021comprehensive,furbank2011phenomics}. Thus, modern sensor technologies have been widely used to achieve non-invasive and high-throughput plant phenotyping\cite{rebetzke2019high}. The most current research shows that optical sensors dominate the detection system \cite{wang2023research,zhou2022intelligent} and various types of two-dimensional (2D) and three-dimensional (3D) imaging systems can directly measure morphological traits of plants, including colors of seeds, leaves, canopies, fruits, and roots, shapes and sizes of seeds, sizes, numbers, areas, textures, angles, architectures, and total volumes of canopies, leaves, and roots, and volumes sizes, shapes, numbers, and spatial distributions of fruits \cite{zhang2018imaging}. 

\begin{figure}[h]
    \centering
    \includegraphics[width=1\linewidth]{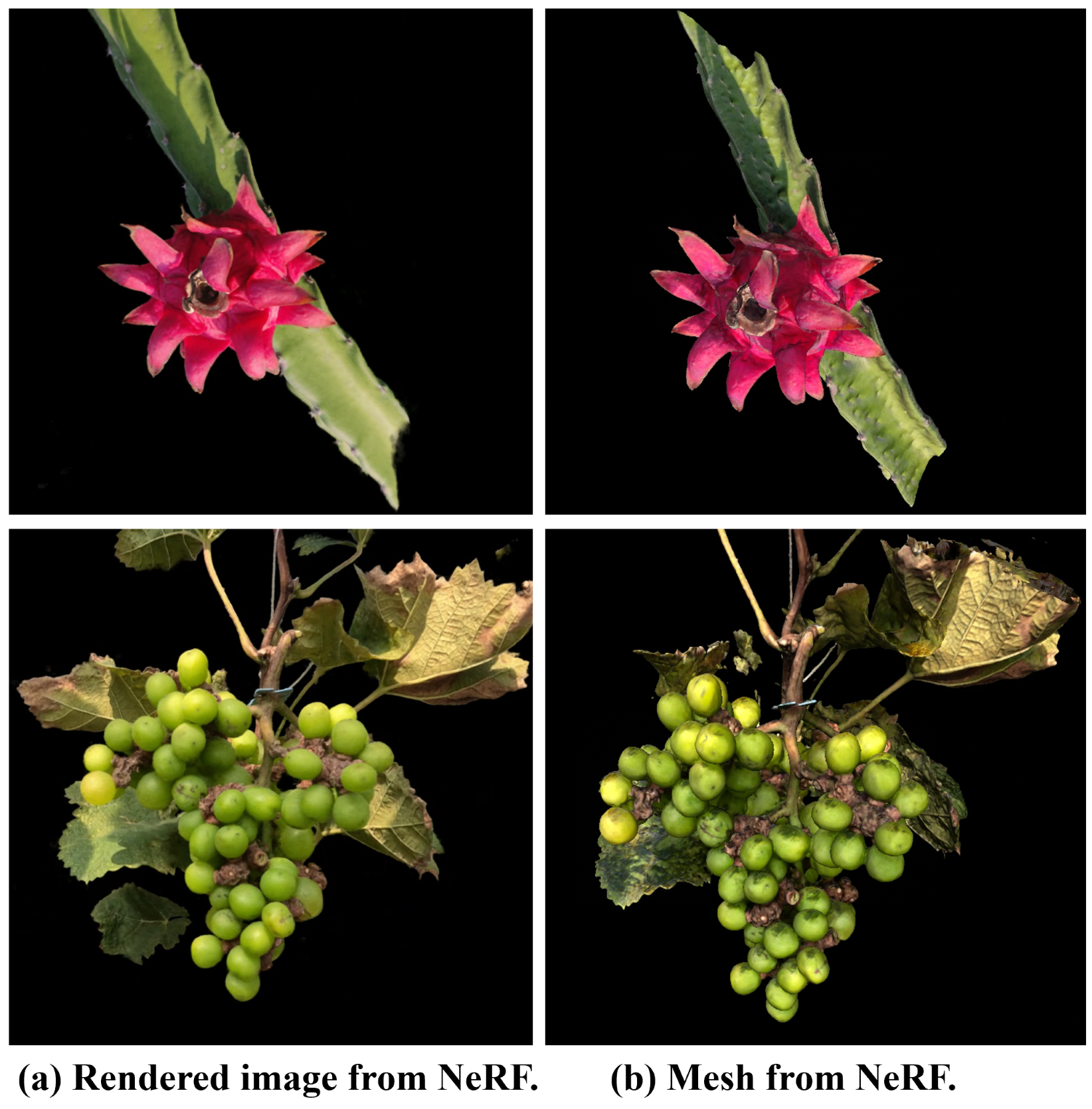}
    \caption{High-fidelity 2D imaging (a) and 3D imaging (b) plant phenotypes from NeRF.}
    \label{fig:enter-label}
\end{figure}

Despite the fact that 2D imaging systems deploy red, green, blue (RGB) camera to measure morphological traits (color, shape, size, and texture) of plants at affordable prices, these methods are limited by the dimensionality of the data and therefore cannot express the geometric form of the plant\cite{zhang2018imaging}. 3D imaging systems enables tracking exact geometry and measurement of plant traits like plant height, plant width, root volume, root surface area, leaf size, leaf width, stem angle and projected canopy area. As a result, the 3D imaging system can keep track of the actual growth status of the plant at the organ level, which is almost impossible for the 2D imaging system\cite{zhang2018imaging,feng2021comprehensive}. However, the current phenotyping methods still face the following challenges: 1) Existing phenotyping techniques cannot obtain multiple types of representations in a single collection paradigm, leading to incomplete data acquisition process. 2) In addition to the RGB camera, some other sensors that are used in phenotyping require a stable operational platform for manual collection. This means that it is difficult to collect phenotypic data beyond RGB image data using robots to replace manual collection. As a result, there's a significant risk of human error. 3) Some sensors compromise data quality to increase flexibility. For instance, while inexpensive depth cameras can substitute for Light Detection And Ranging (LiDAR) to capture point clouds, the resulting data often has lower resolution and is severely degenerated by noise and outliers due to the uniqueness of agricultural environments.

Very recently, a deep learning-based method has the potential to achieve geometric information extraction for 3D imaging while using inexpensive RGB cameras to capture 2D image data of a scene and it has gained widespread attention due to its high-fidelity reconstruction of complex objects and scenes: Neural radiance field (NeRF) \cite{mildenhall2021nerf}. NeRF was first proposed to use volume rendering formula to achieve highly photorealistic view synthesis with implicit neural scene representation via Multi-Layer Perceptron (MLP). Essentially, NeRF uses an MLP network $H_{\Theta}$ to describe the mapping (\ref{Eq:MLP})  between density $\sigma $ and directional emitted color $c=(r,g,b)$ of each point in a 3D scene with the spatial coordinates $(x,y,z) $ and corresponding viewing direction vector $\mathbf {d}$.
\begin{equation}
    \label{Eq:MLP}
    H_{\Theta}(x,y,z,\mathbf{d})\rightarrow(c,\sigma).
\end{equation}

A significant advantage of NeRF is the ability to generate high-quality images of new views that are not merely interpolated, but are true inferences of the underlying scene geometry. This capability is valuable for plant phenotyping, where it is neither feasible nor efficient to manually capture all possible views of a plant or crop. Besides, although the main function of NeRF is implicit scene representation and view synthesis, its density information is stored in the MLP , which provides an important insight and basis for the extraction of geometry. Based on these considerations, research on NeRF is likely to be a key bridge between 2D imaging and 3D imaging, two types of plant phenotyping acquisitions, in order to establish a low-cost, high-throughput, non-invasive plant phenotyping system.

Therefore, this study investigates the performance of the latest NeRF model in 2D view synthesis as well as 3D geometry extraction based on the research content of traditional 2D imaging system and 3D imaging system by acquiring images in a variety of plant growth environments. Specifically, the contributions of this paper are as follows:

\begin{itemize}

\item A novel technique, NeRF, was exported to agricultural applications in this study, particularly for high-fidelity plant phenotyping. 

\item A thorough investigation was conducted on the central tasks of extracting high-fidelity multi-view RGB images and intricate topological geometries using NeRF in actual agricultural scenarios. 

\item A comparison of several state-of-the-art (SOTA) NeRF models in terms of generating new viewpoints and extracting geometric structures was provided, offering invaluable insights for subsequent research.

\end{itemize}

The rest of this paper is organised as follows. Section \ref{section: review}  surveys related work. Section \ref{section: problem statement} delineates the fundamental implementation of rendering novel perspective images and extracting essential geometry. Section \ref{section:method}provides a detailed description of the actual implementation of our methodology. The experiment results and discussion are presented in Section \ref{section:experiment}, followed by the conclusion in Section \ref{section:conclusion}.

\section{Related Works} \label{section: review}
\subsection{Plant phenotyping}

Measuring and analysing plant phenotypes can be used to establish predictive models to assess plant growth characteristics, which are important for precision agriculture as a decision-making tool\cite{feng2021comprehensive}. Therefore, it is crucial to investigate non-invasive, affordable and efficient methods for plant phenotyping\cite{zhao2023phenotyping}. In recent years, a large number of scholars have made many attempts to combine novel sensors with computer technology. Among them, 2D imaging, which studies plant traits such as color through multi-view RGB imaging, and 3D imaging\cite{kolhar2023plant,zhang2018imaging}, which focuses on geometry extraction, have become one of the most important research interests in the field because they serve the most fundamental and widely concerned morphological plant traits. 

\subsection{2D imaging: Multi-view RGB Imaging}
RGB imaging from various perspectives serves distinct purposes in plant phenotyping and growth monitoring\cite{kumar2019image,ubbens2018use}. Kang et al. detected the location of fruits by processing RGB images of apple trees through deep learning \cite{kang2020fast}. Top-view RGB imaging systems are typically employed when examining rosette plants to extract growth rate data. These systems capture top-down RGB images of plants such as Arabidopsis (Arabidopsis thaliana) and tobacco (Nicotiana tabacum) to investigate growth rates under conditions of drought stress, chilling stress, and biotic stress\cite{jansen2009simultaneous,clauw2015leaf}. Plant growth analysis based on top-view images is impacted by challenges such as overlapping leaves and the nastic movement of foliage\cite{dellen2015growth}. These obstacles become particularly pronounced when imaging is limited to a single perspective. Multiview RGB images of cereals, including barley, wheat, rice, sorghum, and various pea field cultivars, are harnessed for the study of growth rates under conditions of drought stress, salt stress, cold stress, and nutrient deficiency\cite{humplik2015automated}. 

\begin{figure*}[h]
    \centering
    \includegraphics[width=0.75\linewidth]{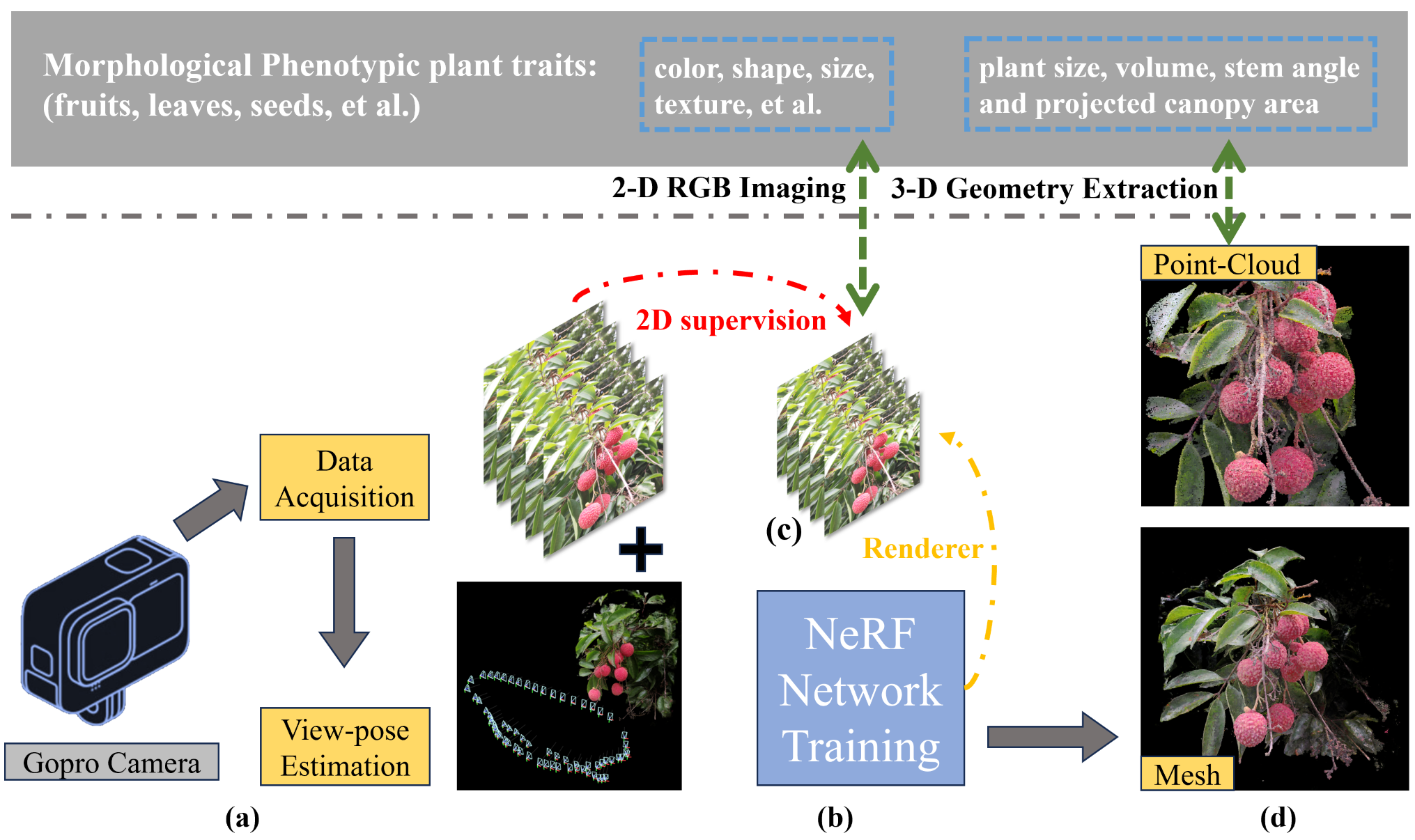}
    \caption{Framework of phenotyping system via NeRF: (a) Data Preparation, (b) Network Training, (c) Images Rendering, (d) Geometry Extraction.  }
    \label{fig:sys-overview}
\end{figure*}

\subsection{3D imaging: Geometry Extraction}
Geometry Extraction of complex unstructured agricultural scenes is the key prerequisite for quantitative extraction of plant metrics. A number of papers applied technologies, which can be divided into two main categories, to obtain the geometric representation of a scene \cite{paulus2019measuring,kang2022uncertainty}. The first is the explicit method, represented by LiDAR, while the other is the implicit method, with Signed Distance Function (SDF) as its representative. 

\textbf{\textit{1) Explicit methods} :} Guo et al. utilized the Realsense D435i to capture continuous multi-view images of the cabbage and input the images into professional 3D reconstruction software called RealityCapture to create 3D point cloud data for calculating the target cabbage dimensions \cite{guo2023improved}. Wu et al.  developed a detachable and adjustable according to the size of the target shoot to aquire multiview stereo (MVS) images and reconstructed 3D poind clouds using MVS-based commercial software \cite{wu2020mvs}. The aforementioned studies calculate the internal parameters of the images, along with the external parameters between them, using feature matching in a series of unordered images. They then proceed to sequentially perform sparse point cloud reconstruction and dense point cloud generation. The quality of the results obtained through these methods is heavily reliant on the resolution and volume of image data, making the process time-consuming. For instance, Guo et al. required 5-8 minutes to capture 150 photographs and an additional 20 minutes or more to complete the reconstruction of a single cabbage. Kang et al. proposed a LiDAR-color fusion-based visual sensing and perception strategy for achieving precise scene comprehension and fruit localization in orchards 
 \cite{kang2022accurate, kang2023semantic}. While this method enhances the density of point cloud data and depth sensor accuracy, it remains costly and time-consuming to accumulate a sufficient number of point clouds. Eugene Kok et al. processed RGB data and depth information from a depth camera using a semantic segmentation network and deep learning skeletonization method to reconstruct spatial information of both visible and hidden branches from a single-view image  \cite{kok2023obscured}. However, the algorithm can only reconstruct trees from a single viewpoint and is not suitable for trees with more complex geometries. Yang et al. developed a system for rapid 3D model reconstruction using RGB-D cameras and the point cloud self-registration method  \cite{yang20223d}. Although they introduced a rotating table to obtain a complete point cloud, this method is only applicable to potted plants and not suitable for field conditions. 
 
\textbf{\textit{3) Implicit methods }:}
Since the conventional reconstruction method represented a 3D scene explicitly using grids of voxels, point clouds, or meshes, the reconstructed 3D shapes were discrete and at a limited resolution. The novel implicit methods parameterize different kinds of features from the scene (for instance, density, color, occupancy probability, SDF value) as a continuous function approximated via an MLP network. Due to the high accuracy of MLP's function fitting, implicit representations of the scene are often accurate at arbitrary resolutions\cite{samavati2023deep}. IM-NET (Chen et al. 2019) trains the network through deep learning using VAE+GAN, which replaced the traditional reconstruction method with a new implicit surface function decoder during the input single view implementation of 3D modeling, resulting in improved reconstruction effectiveness and efficiency\cite{chen2019learning}. Occupancy Networks (Mescheder et al. 2019)  predict binary occupancy rates by acquiring feature vectors and points in space, so that the Occupancy Networks were able to implicitly represent 3D surfaces as continuous decision boundaries for deep neural network classifiers, enabling efficient 3D structural coding through the use of continuous functions to model objects in space\cite{mescheder2019occupancy}. Unlike the principle of Occupancy Networks, DeepSDF (Park et al.) implicitly represents continuous 3D spatial surfaces by directly regressing the Signed Distance Function (SDF). DeepSDF can represent more complex shapes without discrete errors and requires significantly less memory. This concept offers a promising avenue for further research on using neural networks to define implicit scene representations\cite{park2019deepsdf}. Nevertheless, these implicit representations necessitate supervised learning with the pre-existing knowledge of a 3D object's shape, rendering it challenging to directly employ these techniques in real-world environments.

\section{Problem Statement} \label{section: problem statement}
Scene representation on volume rendering is the essential for NeRF to generate novel viewpoint images and geometry. This section gives the formulations of the NeRF working mechanism in both novel-view rendering and geometry reconstruction.

\subsection{Rendering Novel Viewpoints with NeRF}
The volume rendering technique based on ray tracing\cite{bp1984ray} is used to render novel viewpoint images from an invisible MLP. During the rendering process, a ray $r(t)=o+t \mathbf{d}$ is emitted into the 3D scene from a given camera position $o=(x_o,y_o,z_o)$, where $t$ is the straight-line distance from the point on the ray to the camera's origin $o$ and $\mathbf{d}$ is the 3D Cartesian unit vector representing viewing direction. As the ray travels, a sufficient number of spatial points of this ray are sampled, and the $\sigma$ and $c$ of each point are queried to the $H_{\Theta}$. Finally, the equation (\ref{Eq:volume rendering}) from classical volume rendering \cite{bp1984ray} is used to accumulate all the sampled points and obtain the color value of the corresponding pixel on the image plane.
\begin{equation}
    \label{Eq:volume rendering}
    C(\mathbf{r})=\int_{t_1}^{t_2}T(t)\cdot\sigma(\mathbf{r}(t))\cdot\mathbf{c}(\mathbf{r}(t),\mathbf{d})\cdot dt,
\end{equation}
where $T(t)$ denotes the transmittance function, alternatively referred to as the accumulated density. This function quantifies the possibility that a ray traverses the distance from $t_1$ to $t_2$ without runing into an obstruction, as described by the following equation: 
\begin{equation}
    \label{Eq:Tt}
    T(t)=\exp(-\int_{t_1}^t\sigma(\mathbf{r}(u))\cdot du),
\end{equation}

For every pixel, a squared error photometric loss is employed for the optimization of the wight $\Theta$ of MLP . When applied across the entire image, this loss is represented as follows: 
\begin{equation}
    \label{Eq:L}
   \mathcal{L}_{\mathrm{NeRF}}=\sum_{r\in R}||C(\mathbf{r})-C_{gt}(\mathbf{r})||^2 .
\end{equation}\\
where $C_{gt}(\mathbf{r})$ is the ground truth color of the training image pixels associated with the ray $\mathbf{r}$, and $R$ is a batch of rays associated with the image to be synthesised.

\subsection{Extraction Geometry From NeRF } \label{section:nerf2mesh} 
\subsubsection{Point-Cloud Extraction from NeRF}

While NeRF primarily focuses on the reconstruction and rendering of scenes from novel viewpoints, it inherently contains a wealth of 3D structural information, making it possible to extract point-cloud data. The continuous feature of NeRF's representation allows the inference of spatial geometries by observing changes in radiance and density along camera rays.

The crucial step in point-cloud extraction is depth estimation. By analyzing the transmittance function \( T(t) \) in (\ref{Eq:Tt}) along a ray \( \mathbf{r}(t) \), one can observe the depth where the function experiences significant change, indicating the presence of a surface. The depth at which \( T(t) \) sees a sharp decline is usually aligned with the surface of an object within the scene. Mathematically, this depth \( t_{surface} \) for a ray \( \mathbf{r}(t) \) can be pinpointed as the position where the transmittance's rate of change is most abrupt by minimizing the first order derivative of  \( T(t) \) relative to $t$:
\begin{equation}
    t_{surface} \approx \arg\min_t \left( \frac{dT(t)}{dt} \right),
\end{equation}
With the depth approximated, the next step is to calculate the corresponding 3D point on $r(t)=o+t \mathbf{d}$ :
\begin{equation}
    \mathbf{p} = \mathbf{o} + t_{surface} \mathbf{d},
\end{equation}
Where \( \mathbf{p} \) represents the 3D point, \( \mathbf{o} \) is the camera's origin, \( \mathbf{d} \) signifies the ray direction and $t_{surface}$ t represents the distance travelled by the ray as it crosses the surface.

Upon determining the 3D position, the color value at this position can be directly queried from NeRF (\ref{Eq:MLP}):
\begin{equation}
    \mathbf{color} = \mathbf{c}(\mathbf{p}, \mathbf{d}).
\end{equation}

Repeating these steps for every pixel across one or multiple images generates a dense point-cloud. Each point within this cloud corresponds to a surface in the original scene, carrying a color that mirrors the appearance of that surface under the sampled viewing direction.
\subsubsection{Mesh Extraction from NeRF} 
Given a predefined 3D region of interest, a set of spatial points $P=\{p_1,p_2,...,p_n\}$ is generated via dense volumetric sampling. For each point $p_i\in P$ , it's evaluated through the NeRF model to obtain The density values, $\sigma(p_i)=\mathrm{NeRF}_\sigma(p_i)$, form the basis for surface extraction.
The Marching Cubes \cite{lorensen1998marching} algorithm identifies the isosurface by thresholding the density values: 
\begin{equation}
    M=\text{MarchingCubes}(P,\sigma_{\text{threshold}}) ,
\end{equation}
Where $M$ is the resultant mesh and $\sigma_{threshold}$ is an optimal density value demarcating the object's boundary.  

For every vertex $v_j$ in mesh $M$ , a viewing ray $r_j$ is constructed and queried  $ \mathbf{c}(v_j)=\mathrm{NeRF}_\mathbf{c}(v_j,r_j)$ in NeRF. Those radiance values derived from NeRF are mapped onto mesh $M$, assigning color to each vertex: 
\begin{equation}
    \operatorname{Color}(v_j)=\mathbf{c}(v_j)  ,
\end{equation}

For a 2D texture representation, the vertex-colored mesh undergoes UV unwrapping \cite{sander2001texture}. To minimize distortion, algorithms like Least Squares Conformal Mapping (LSCM) \cite{levy2023least} can be employed. The objective of LSCM is to minimize the conformal energy: 
\begin{equation}
    E(u,v)=\int_{\Omega}\left(|\nabla u|^2+|\nabla v|^2\right)dA.
\end{equation}
Where $(u,v)$ are the 2D texture coordinates for each vertex in $M$, $\Omega$ represents the object's surface, and $dA$ is a differential area element on the mesh's surface, indicating that the energy is computed by integrating over the entire surface of the mesh. 

\section{Methods} \label{section:method}
To adapt NeRF effectively for agricultural scenarios, we made specific enhancements to address its initial limitations. Recognizing that the original NeRF training was slow, we introduced hash encoding of Instant-ngp\cite{muller2022instant} to speed up the training process. To strengthen the geometric constraints within the model, we brought in a rendering method based on SDF\cite{wang2021neus,zhao2022human}. In this chapter, we will walk through the entire process, from image data collection and preparation for training to the architecture and training of the network model. Finally, we will give an overview of the experimental setup used in our study.

\subsection{Data Acquisition} \label{subsection:data aquisition}
\begin{figure}[ht]
    \centering
    \includegraphics[width=1\linewidth]{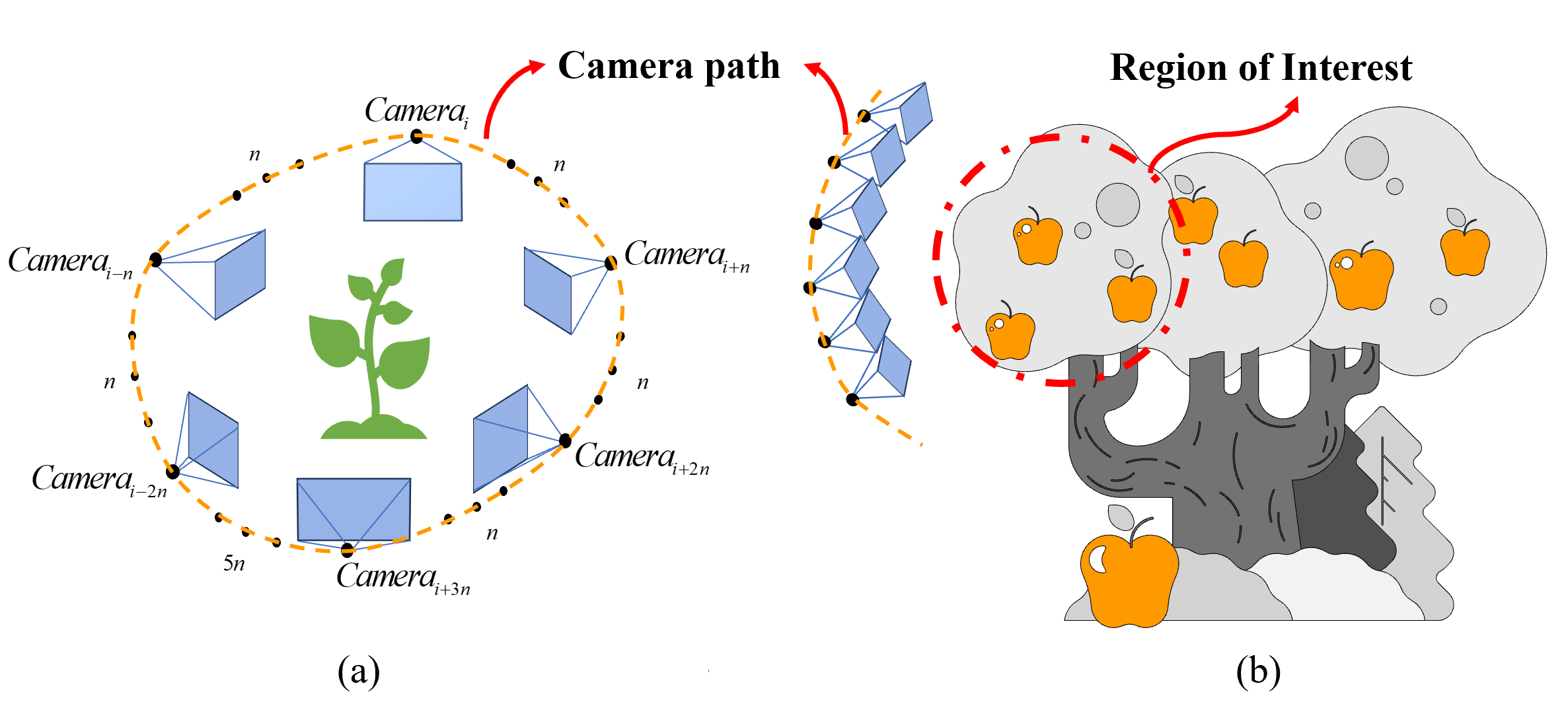}
    \caption{(a) 360° image capturing, (b) front views capturing.}
    \label{fig:data-aqure}
\end{figure}
We captured high-resolution plant images from various agricultural scenes using the GoPro Hero 11 action camera in Tab. \ref{tab:camera_parameters}. To reduce motion blur and graphic quality issues, the camera was set to work at 120 Hz in a 4K resolution linear imaging mode. This setting allowed us to collect image data at a rate of 120 frames per second with a resolution of 3840x2160 pixels. For single plants in Fig. \ref{fig:data-aqure}(a), we aimed to capture 360° all-around images to cover all details. For larger, more complex scenes in Fig. \ref{fig:data-aqure}(b), we chose front views of regions of interest and took images from multiple angles. 
\begin{table}[h]
    \centering
    \begin{tabular}{cc}
    \toprule
    \textbf{Parameter} & \textbf{Value} \\
    \midrule
    Name & GoProHERO11 \\
    Weight & 470.00g \\
    Resolution & 4K+ \\
    Lens Stabilization & Electronic \\
    Battery Life & 90 minutes \\
    \bottomrule
    \end{tabular}
    \caption{Camera Parameters for GoProHERO11}
    \label{tab:camera_parameters}
\end{table}

\subsection{View-pose estimation} \label{section:colmap}
COLMAP\cite{schonberger2016structure}, a SOTA Structure-from-Motion (SfM) and Multi-View Stereo (MVS) pipeline, reconstructs 3D models from unordered image sets. The pipeline of SfM was used to estimate images` poses as as a prior to supervise the training of the network. 

For every image in the datasets, COLMAP detects and describes local features. A pairwise matching algorithm then associates keypoints based on their descriptors. Formally, if $p_i$ and $p_j$ are keypoints in images $I_i$ and $I_j$ respectively, their match is established based on: 
\begin{equation}
    D(p_i,p_j)=\|\mathrm{desc}(p_i)-\mathrm{desc}(p_j)\|_2 ,
\end{equation}
Where $\mathrm{desc}(p)$ returns the descriptor for keypoint $p$. 
After feature matching, geometrically consistent matches are pinpointed by employing a fundamental or essential matrix. The essential matrix, denoted as $E$, captures the geometric relationship between two calibrated images. It is a matrix that relates corresponding points in one image to epipolar lines in the other image. Formally:
\begin{equation}
    p_j^TEp_i=0 ,
\end{equation}
Where $p_i$ and $p_j$ are corresponding points in homogeneous coordinates. 

COLMAP utilizes an incremental approach to SfM. Starting with a pair of images with the largest number of geometrically consistent matches, the scene is incrementally expanded by registering additional images based on shared keypoints. 

After initial camera pose estimation, COLMAP refines the camera parameters, 3D structure, and even image keypoints simultaneously using bundle adjustment. The objective function $J$ being minimized is: 
\begin{equation}
    J=\sum_{i,j}w_{ij}\|p_j-\pi(P_i,X_j)\|^2 .
\end{equation}
Where $w_{ij}$ is a visibility term, which is 1 if point $X_i$ is visible in image $I_i$ and 0 otherwise. $\pi$ is the projection function. $P_i$ is the projection matrix of the i-th image. $X_j$ is the 3D position of the j-th point. 

\subsection{Learning From NeRF}

\begin{figure*}[ht]
    \centering
    \includegraphics[width=0.85\linewidth]{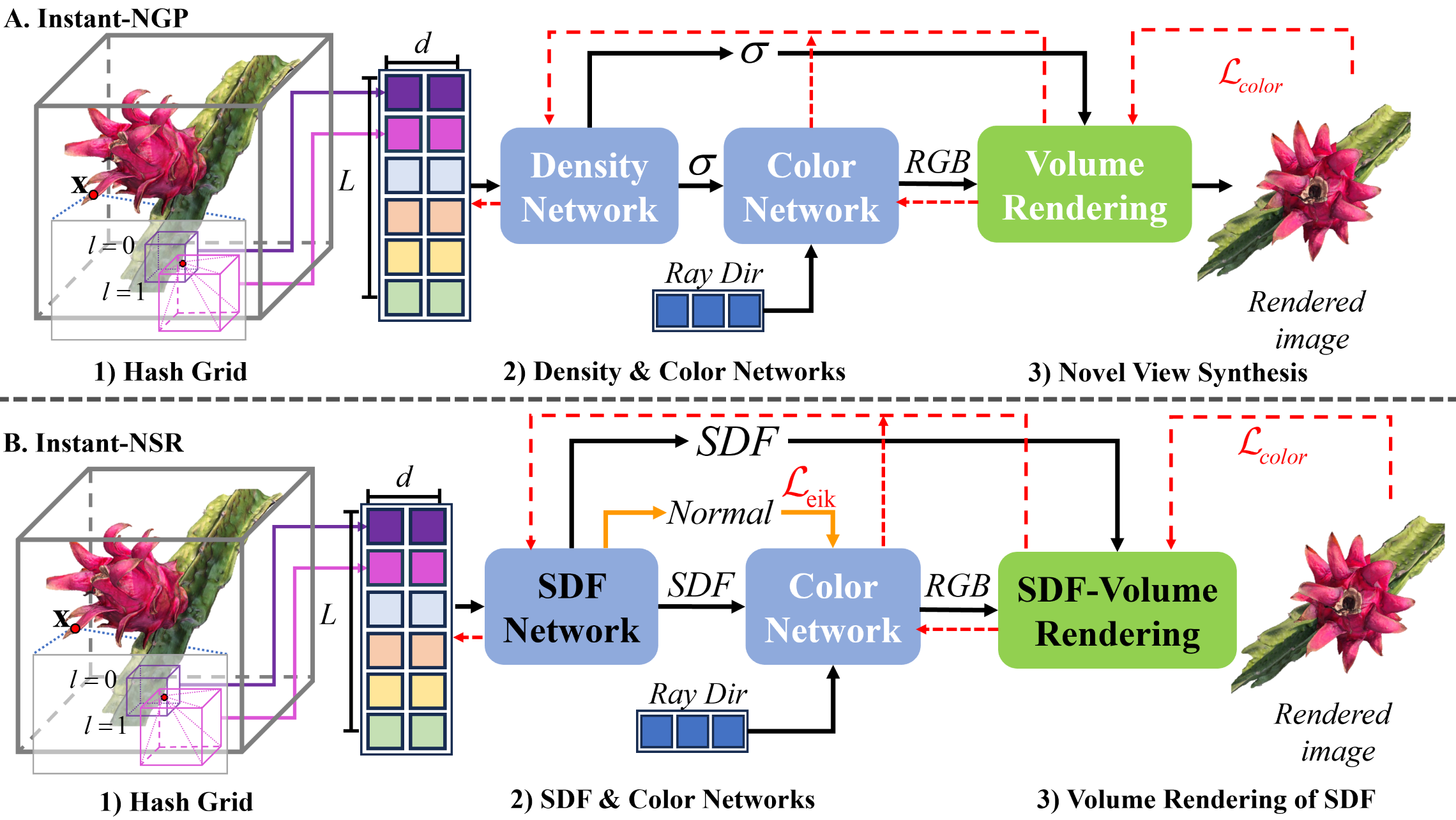}
    \caption{Pipeline of Instant-NGP and Instant-NSR. \textbf{A). Instant-NGP: }Given a 3D point $\mathbf{x}$, The 1)hash grid corresponding to each level \textit{l} in the voxel grid is interpolated to hash encoding, then the density and color values are predicted by the 2)MLPs of density and color, and the color of the pixel is calculated by 3)volumetric rendering.  \textbf{B). Instant-NSR:} Compared to the previous\textbf{NGP}, both the 2) MLPs and the 3)volume rendering are based on SDF and employ an extra normal regularisation to strengthen the geometrical constraints in the network training.}
    \label{fig:pipline}
\end{figure*}
\textbf{Multi-resolution hash encoding.} To address the drawback of slow nerf model training, Instant-NGP \cite{muller2022instant} makes use of a multi-resolution hashed positional encoding as additional learned features, the model could represent scenes accurately with tiny and efficient MLPs. In detail, Instant-NGP operates on the premise that the object to be reconstructed is enclosed within multi-resolution voxel grids. Each of these voxel grids at different resolutions is then correspondingly linked to a hash table, featuring a fixed-size array of adaptable feature vectors.

For any spatial point $\mathbf{x}\in\mathbb{R}^{3}$ within various resolution grids, it obtains the hash encoding $h^i(\mathbf{x})\in\mathbb{R}^d$ (d is the dimension of a feature vector, $i=1,...,L$) corresponding to the respective level by trilinear interpolation. The hash encodings at all $L$ levels are subsequently concatenated to form the multi-resolution hash encoding $h(\mathbf{x})=\{h^i(\mathbf{x})\}_{i=1}^L\in \mathbb{R}^{L\times d}$ .

\textbf{Volume rendering of SDF.}  To precisely extract the geometric surface from NeRF's implicit representation, Neus \cite{wang2021neus} proposes to represent 3D scene as a signed distance function (SDF) $\begin{aligned}f(\mathbf{x}):\mathbb{R}^3\to\mathbb{R}\end{aligned}$ instead of NeRF`s density field and introduce a \textit{\textbf{S}-density} $\phi_b(f(\mathbf{x}))=\begin{aligned}be^{-bf(\mathbf{x})}/(1+e^{-bf(\mathbf{x})})^2\end{aligned}$, where $b$ is a trainable hyper parameter and gradually increases to a large number as the network training converges. And the surface $\mathcal{S}$ can be extracted by the zero-level set of its SDF:
\begin{equation}
    \label{Eq:SDFs}
    \mathcal{S}=\left\{\mathrm{x}\in\mathbb{R}^3|f(\mathbf{x})=0\right\},
\end{equation}
To train the neural SDF representation, Neus followed NeRF`s volume rendering equation \ref{Eq:volume rendering}. Given a pixel, the renderer emitted a ray from this pixel as $\{\mathbf{p}(t)=\mathbf{o}+t\mathbf{v}|t\geq0\}$, where $\mathbf{o}$ is origin of the ray and $\mathbf{v}$ is the ray direction and accumulate the colors along the ray by:
\begin{equation}
    \label{Eq:SDFrender}
    C(\mathbf{o},\mathbf{v})=\int_0^{+\infty}w(t)c(\mathbf{p}(t),\mathbf{v})\mathrm{d}t,
\end{equation}
where $C(\mathbf{o},\mathbf{v})$ is the rendered color for this pixel, and $c(\mathbf{p}(t),\mathbf{v})$ the sampled colors along the ray. Especially, the weight $w(t)$ for point $\mathbf{p}(t)$ is rebuilt by unbiased and occlusion-aware properties to guarantee that the surface of an actual object contributes the most to the rendering result, that is:
\begin{equation}
    \label{Eq:wtneus}
    w(t)=\frac{\phi_s(f(\mathbf{p}(t)))}{\int_0^{+\infty}\phi_s(f(\mathbf{p}(u)))\mathrm{d}u}.
\end{equation}

\textbf{Truncated SDF Hash Grids.}  To increase the stability of network training, here, we introduce a neural surface reconstruction method that accelerates training with Hash encoding, Instant-NSR\cite{zhao2022human}. as in Fig. \ref{fig:pipline}, this method is similar to Instant-NGP \cite{muller2022instant}in that it Hash encodes points in spatial at the front-end of the neural network, but employs Neus's SDF architecture \cite{wang2021neus} in the neural network instead of the NGP's density architecture, and in the image renderer, also SDF-based volume rendering formulation (\ref{Eq:SDFrender}) is used. 

In addition, Instant-NSR uses Truncated SDF (TSDF) to skilfully solve the convergence problem caused by applying SDF representations to hash coding frameworks. Since original SDF-based methods utilize cumulative density distribution $\phi_b(f(\mathbf{x}))=\begin{aligned}be^{-bf(\mathbf{x})}/(1+e^{-bf(\mathbf{x})})^2\end{aligned}$ to the compute $w(t)$ in equation (\ref{Eq:wtneus}), the term $-bf(\mathbf{x})$ will be a large positive number when $b$ is increased, resulting in $e^{-bf(\mathbf{x})}$ closing to infinity.  This numerical instability will cause the network to converge hardly during the training process. The characteristic of TSDF value between -1 to 1 can effectively prevent the occurrence of network divergence caused by numerical overflow. Therefore, we utilize the sigmoid function $\pi(\cdot)$ after the SDF output of the network to achieve the truncation effect of the TSDF, as below:
\begin{equation}
    \pi(f(\mathbf{x}))=\frac{1-e^{-bf(\mathbf{x})}}{1+e^{-bf(\mathbf{x})}}.
\end{equation}
Thus, we can now replace the formula $\phi_b(f(\mathbf{x}))=\begin{aligned}be^{-bf(\mathbf{x})}/(1+e^{-bf(\mathbf{x})})^2\end{aligned}$ in Neus with $\phi_b(f(\mathbf{x}))=\begin{aligned}be^{-b\pi(f(\mathbf{x}))}/(1+e^{-b\pi(f(\mathbf{x}))})^2\end{aligned}$.

\subsection{Network Training}

In order to obtain an optimal representation of the scene via our neural network model, we employ a compound loss function. This loss function is constructed using two primary components: the rendering loss and the eikonal loss.

\subsubsection{Rendering Loss}

The primary goal of our methd is to produce high-quality renderings that closely match the ground truth images. Therefore, the rendering loss is crucial as it quantifies the discrepancy between the rendered images from the network and the actual images, and we generally apply this ${\mathcal{L}_{color}}$ in both Instant-NGP and Instant-NSR.

Given a set of ground truth images \(I_{gt}\) and the corresponding set of images rendered by the network \(I_{pred}\), the rendering loss ${\mathcal{L}_{color}} $ is defined as:

\begin{equation}
{\mathcal{L}_{color}} = \frac{1}{N} \sum_{i=1}^{N} \| I_{gt}^{(i)} - I_{pred}^{(i)} \|_2^2 ,
\end{equation}

where \(N\) is the total number of images in the datasets.

\subsubsection{Eikonal Loss}

While the rendering loss ${\mathcal{L}_{color}}$ ensures the visual accuracy of the rendered images, the eikonal loss ${\mathcal{L}_{eik}}$ is used in Instant-NSR to ensure that the estimated SDF values conform to the properties of a true SDF following one of the primary properties that the gradient $\nabla f(\mathbf{x})$ of the SDF should have a magnitude of $1$ everywhere.

Given an SDF represented by \( f(\mathbf{x}) \), the eikonal loss ${\mathcal{L}_{eik}}$  is given by:

\begin{equation}
{\mathcal{L}_{eik}}= \frac{1}{M} \sum_{j=1}^{M} \left( \|\nabla f(\mathbf{x}_j)\|_2 - 1 \right)^2 ,
\end{equation}

where \(M\) is the total number of sampled points from the scene, and $\nabla f(\mathbf{x}_j)$ is the gradient of the SDF for the j-th point.

\subsubsection{Total Loss}

Combining both losses, the total loss function \( L_{total} \) used to train the Instant-NSR network is:

\begin{equation}
L_{total} = \alpha L_{render} + \beta L_{eik}.
\end{equation}

where \( \alpha \) and \( \beta \) are weighting factors that balance the contribution of the two losses. These weights are hyperparameters and are chosen based on cross-validation to achieve the best performance on a validation set.

\subsection{Implementation details}

\subsubsection{Image Processing and Data Preparation}
After the acquisition, the COLMAP toolbox was utilized to compute the camera parameters and relative poses for the captured images. An essential step in data preparation involved the conversion of the computed image parameters into the Local Light Field Fusion (LLFF) \cite{mildenhall2019local} format. This transformation was achieved through a TensorFlow implementation of the LLFF toolbox. The selection of the LLFF format was based on its robust representation, which is indispensable for the neural network models employed in this study.

\subsubsection{Computing Infrastructure}

All neural network training and computational experiments were conducted on an Ubuntu-based platform. The hardware specifications of the system included an Intel Core i9-13900 CPU. For graphic processing and deep learning computations, the system was equipped with two Colorful GeForce RTX 3090 graphics cards, providing a combined video memory of 48GB. This high-end setup ensured the efficient execution of data-intensive processes and neural network operations.

\section{Experiment and Discussion} \label{section:experiment}

\subsection{Experimental Method}
\begin{figure}[ht]
    \centering
    \includegraphics[width=1\linewidth]{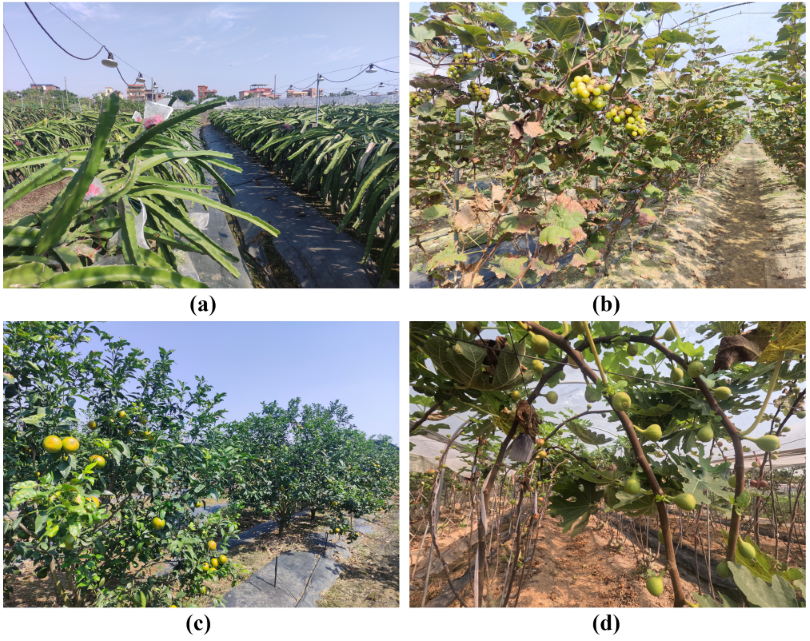}
    \caption{Photographs of indoor and outdoor orchards: 
    (a) pitahaya ochard, (b) grape ochard, (c) orange ochard, (d) fig ochard. }
    \label{fig:enter-label}
\end{figure}
\begin{figure*}[ht]
    \centering
    \includegraphics[width=1\linewidth]{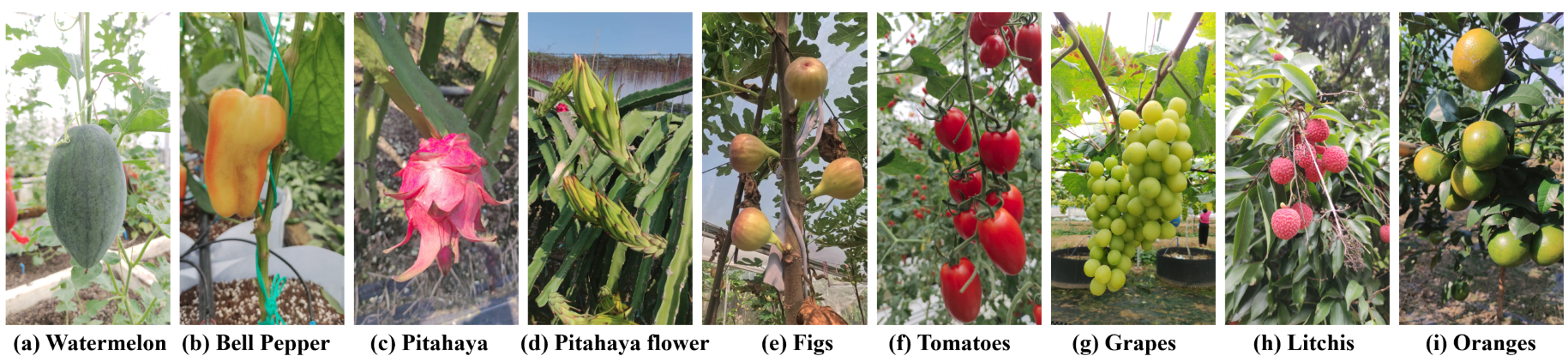}
    \caption{Demonstration of our datasets with three progressive levels: $L_1$: (a),(b),(c); $L_2$: (d),(e),(f); $L_3$: (g),(h),(i).}
    \label{fig:showdata}
\end{figure*}

In this study, we utilised a high-speed motion camera, GoPro Hero 11, to acquire the image datasets for our experiments, which were divided into three levels, $L_1$, $L_2$, and $L_3$, based on the geometrical distribution of important structures such as leaves and fruits of the plants in them. Firstly, we tested the performance of Instant-NGP in rendering the images in these datasets, and secondly, we tested two different geometrically expressed NeRF models, Instant-NGP based on the density field and Instant-NSR based on the SDF, for their ability to extract the geometrical models of the plants in these datasets.

\subsection{Data Preparation}\label{section:data preparation}
In this section, we collected image datasets of litchi at the Litchi Expo Park in Zengcheng District, Guangzhou, image datasets of bell peppers, tomatoes, and watermelons planted in greenhouses at the Baiyun Experimental Base of the Guangzhou Academy of Agricultural Sciences, and image datasets of grapes, pitahaya, pitahaya flowers, oranges, and figs at the Shangguo Ecological Picking Garden in Panyu District, Guangzhou in Fig.\ref{fig:showdata}.
\subsubsection{Levelling of datasets}
Furthermore, this study classified the scenes based on the interplay and occlusion among these key plant constituents. Three distinct levels were defined to represent these datasets: $L_1$, $L_2$, and $L_3$.  

\begin{itemize}
\item  $L_1$ represents scenes where the fruits, leaves and branches are clearly visible with minimal to no occlusion between them. In such scenarios, each component is clearly visible, making it an ideal representation of less dense plant geometry.

\item $L_2$ represents scenes with a slightly denser configuration. Here, several fruits overlap each other and there is slight occlusion by the leaves. This level is moderately challenging and represents environments where components begin to intertwine.

\item $L_3$ is indicative of the most complex and confused scenarios. In these scenes, fruit, leaves and branches are chaotically distributed and the geometric topology is highly complex. Such environments resemble dense plant canopies and thickets, where distinguishing individual components becomes particularly challenging.
\end{itemize}

\subsubsection{Data processing and format conversion}
\begin{figure}
    \centering
    \includegraphics[width=1\linewidth]{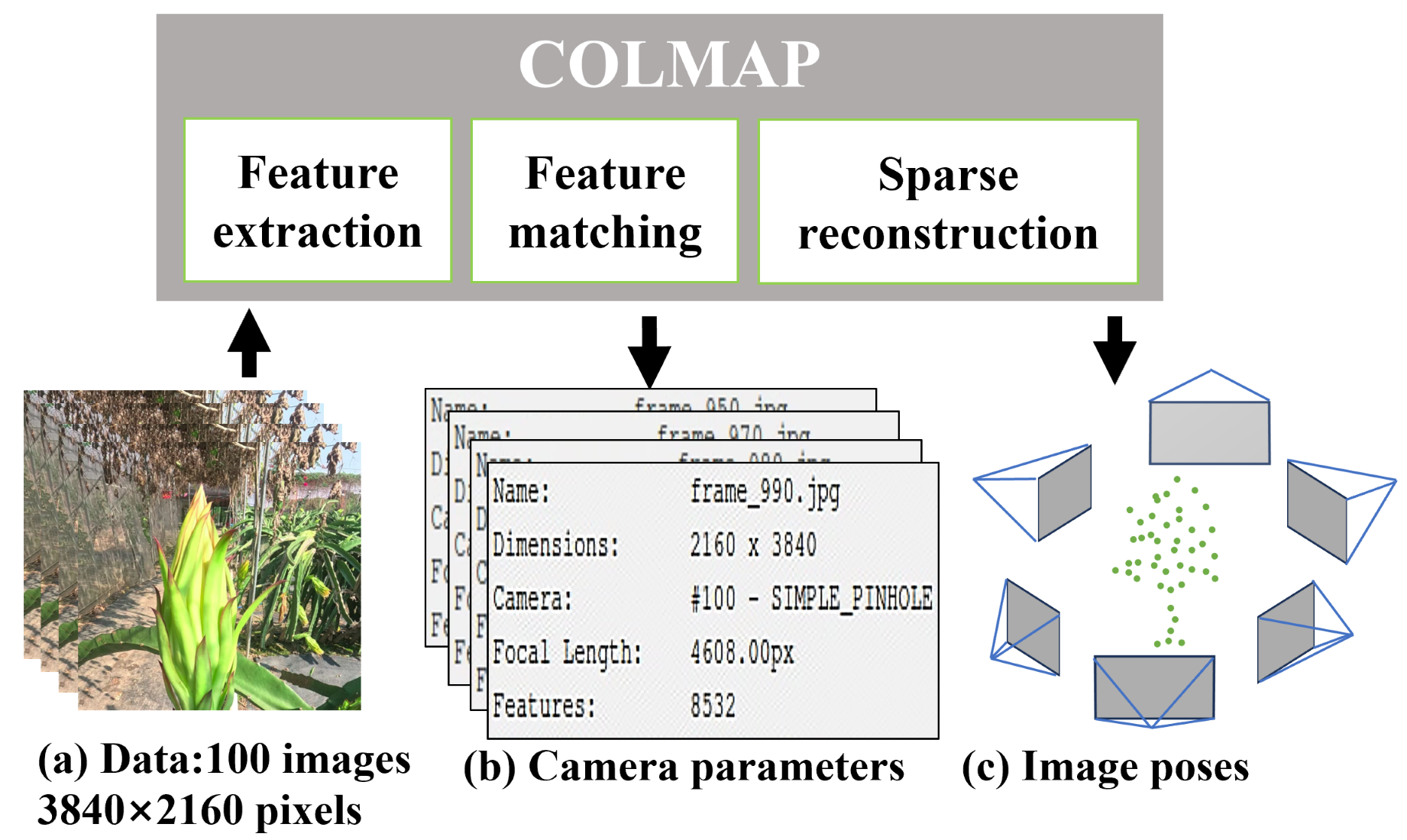}
    \caption{Framework of data processing. }
    \label{fig:data_pro}
\end{figure}
Before training the NeRF models, first of all, we apply COLMAP to compute the camera parameters and the poses between images according to the method in Sec. \ref{section:colmap}. To assist NeRF in processing datasets from different sources, these images, along with their correlated camera poses, camera parameters, are required to be converted into a certain format.

The Local Light Field Fusion (LLFF) format \cite{mildenhall2019local}was designed for capturing real-world scenes using a series of photographs taken from various viewpoints. Distinct from traditional light field cameras, LLFF does not require specialized hardware but leverages conventional cameras. By moving the camera throughout the scene and capturing multiple images, a scene representation is generated. Given NeRF's (Neural Radiance Fields) objective to learn 3D representations of a scene from a series of images, the LLFF datasets format naturally becomes an optimal choice for representing input data from real scenes.

A LLFF datasets typically comprises the following crucial components: (1) A set of images captured from different perspectives, (2) Camera intrinsic parameters, (3) Camera extrinsic parameters. Within the previous steps, we have obtained the (1) image sequences and (2) intrinsic parameters of cameras. To represent the output camera poses of COLMAP in the LLFF format, it is needed to invert the transformation from a world-to-camera format (COLMAP) to camera-to-world format for LLFF.

Specifically, COLMAP outputs a rotation matrix $R$ and a translation vector $t$ for each camera`s pose $C=-R^T\cdot t$. For a rotation matrix, which is orthogonal, the inverse $R^{-1}$ is equal to the transpose $R^{T}$. The translation in the world-to-camera format can be found by transforming the camera's position in the world coordinates using the inverse rotation: 
\begin{equation}
    t^{\prime}=-R^T\cdot t
\end{equation}
Therefore, the LLFF camera-to-world transformation matrix is constructed as:

 \begin{equation}
     M=\begin{bmatrix}R^T&t^{\prime}\\0&1\end{bmatrix}
 \end{equation}

\begin{figure}
    \centering
    \includegraphics[width=1\linewidth]{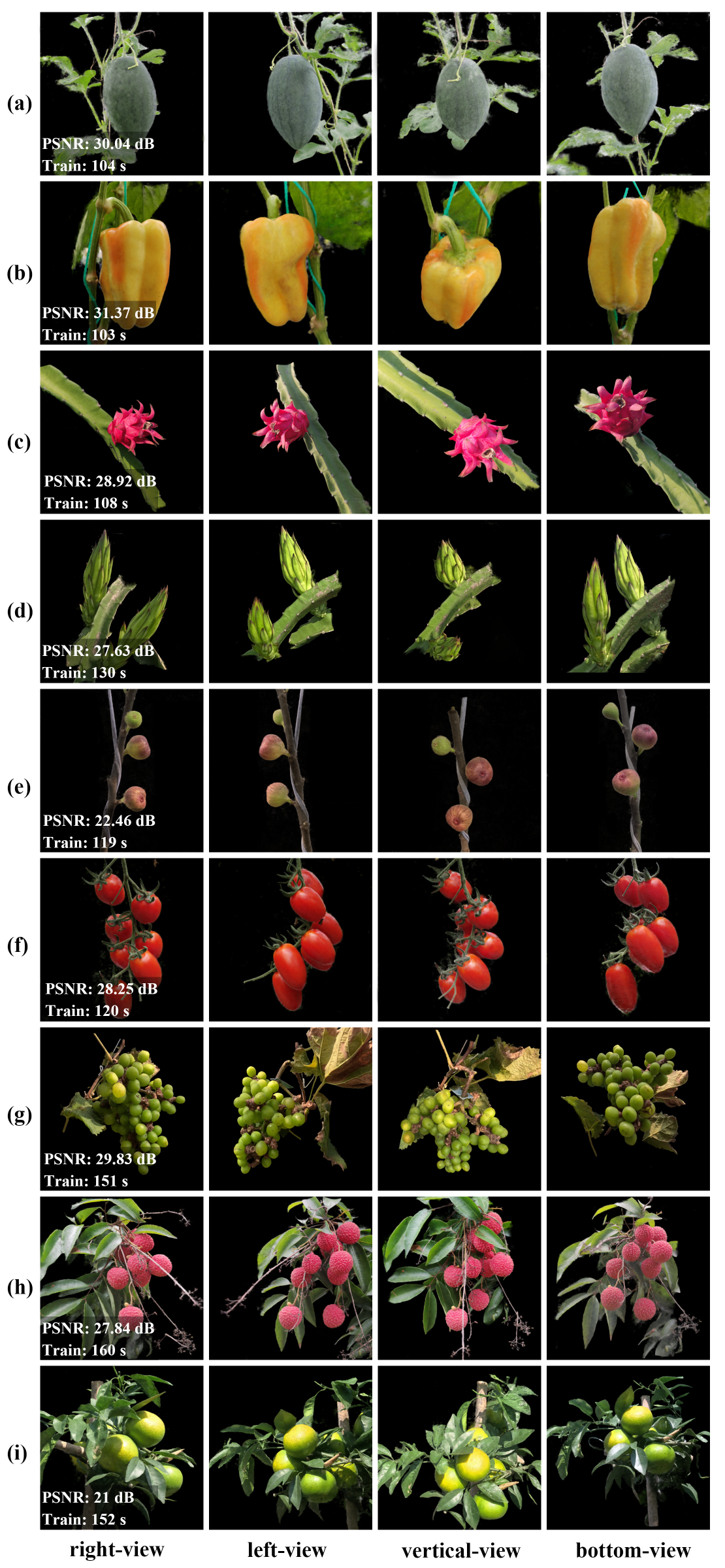}
    \caption{Novel-view image synthesised results (from different views in sequence.) by using Instant-NGP that trained on $L_1$,$L_2$,$L_3$ datasets.}
    \label{fig:rendering results}
\end{figure}

\subsection{Demonstration in 2D imaging of NeRF}   \label{subsection:rendering results}
This section demonstrates that the results of real-time rendering of our datasets in the quickest training NeRF model  Instant-NGP \cite{muller2022instant}. To explore the effect of the complexity of the scene on the NeRF rendering quality, we set the number of images in all datasets to 100 images, all of which were captured by GoPro cameras in 120 HZ, linear imaging mode, with a resolution of 3840x2160 pixels, according to the settings in Sec. \ref{subsection:data aquisition}.

The Peak Signal-to-Noise Ratio (PSNR) has been widely recognized as an essential metric for the quantitative evaluation of image quality. Predominantly used in the domain of image compression and reconstruction, its application has further expanded into the emerging realms of computer vision and neural graphics\cite{mildenhall2021nerf}. Higher PSNR values imply superior image fidelity, indicating a closer match to the reference. The foundation of PSNR lies in the Mean Squared Error (MSE), which quantifies the average squared discrepancies between the pixel values of the reference and the examined images. Formally, for an image of size $M×N$, the MSE is defined as:
\begin{equation}
    \label{Eq:MSE}
    \mathrm{MSE}=\frac1{M\times N}\sum_{i=1}^M\sum_{j=1}^N[I_{\mathrm{reference}}(i,j)-I_{\mathrm{examined}}(i,j)]^2
\end{equation}
Where $I_{\text{reference}}$ and $I_{\text{examined}}$ represent the pixel intensities of the reference and examined images, respectively. With the MSE in hand, the PSNR is calculated using: 
\begin{equation}
    \label{Eq:PSNR}
    \mathrm{PSNR}=10\times\log_{10}\left(\frac{MAX_I^2}{\mathrm{MSE}}\right)
\end{equation}
where $MAX_I$ signifies the maximum feasible pixel intensity for the image. ( For instance, for a typical 8-bit grayscale image, $MAX_I$ equals 255. )

To quantitatively evaluate these experimental results, we referenced the original NeRF paper's real-world dataset benchmark which recorded a PSNR of 26.50 dB in their Real Forward-Facing dataset. On this baseline, a dataset with a PSNR close to or higher than 26.50 dB indicates high reconstruction quality, and the opposite indicates that the dataset struggles to converge well. 

Fig. \ref{fig:rendering results} shows the training time of our full datasets in  Instant-NGP and the PSNR for each scene. To demonstrate the power of NeRF to synthesise new views, we have selected four views other than the training data used for the model, namely right, left, vertical and bottom views.

\subsection{Demonstration in 3D imaging of NeRF}
\begin{figure}
    \centering
    \includegraphics[width=1\linewidth]{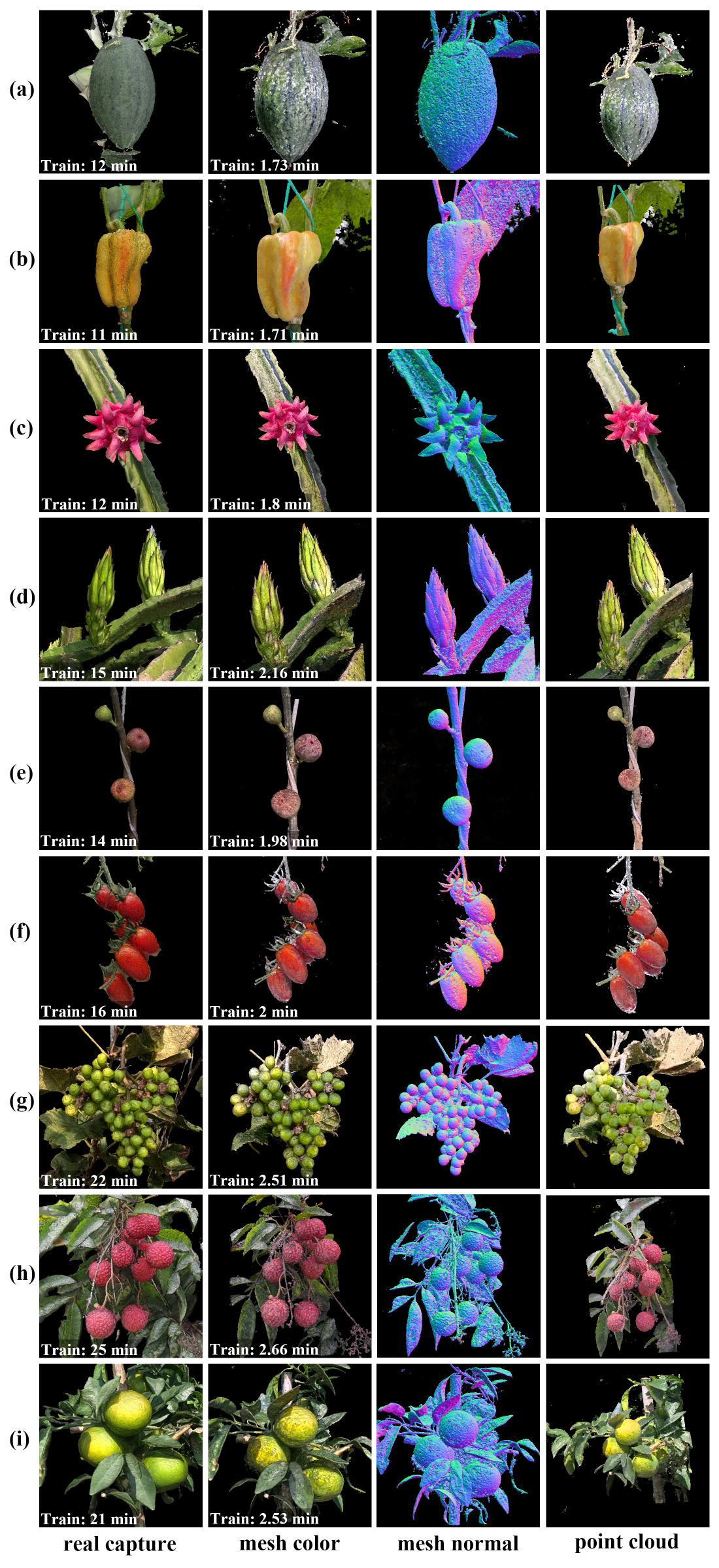}
    \caption{3D models extracted from $L_1$,$L_2$,$L_3$ datasets. The leftmost line of data is the real reference views of the plants, followed by the mesh models extracted from Reality-Capture, the normal mapping models, the textured mesh models, and the point clouds extracted from Instant-NGP in sequence.}
    \label{fig:Geometry-ngp}
\end{figure}

This section details the experimental results of the geometry extraction from NeRF based on our plant datasets. First of all, we extract the point clouds and meshes of the plants from Instant-NGP according to the method introduced in Sec.\ref{section:nerf2mesh}, and demonstrate these results comprehensively in Fig.\ref{fig:Geometry-ngp}. And note that in this experiment, the mesh models generated by Reality capture is used as a reference for the geometric extraction results, because NeRF is supervised by 2D image data without the ground truth of 3D information. (Reality capture is a commercial MVS-based modelling software, which integrates the core methods of MVS as well as comprehensive steps, including photo alignment, feature extraction, feature matching, camera viewpoint calculation, and 3D point cloud reconstruction\cite{wu2020mvs}.)

\begin{figure*}
    \centering
    \includegraphics[width=1\linewidth]{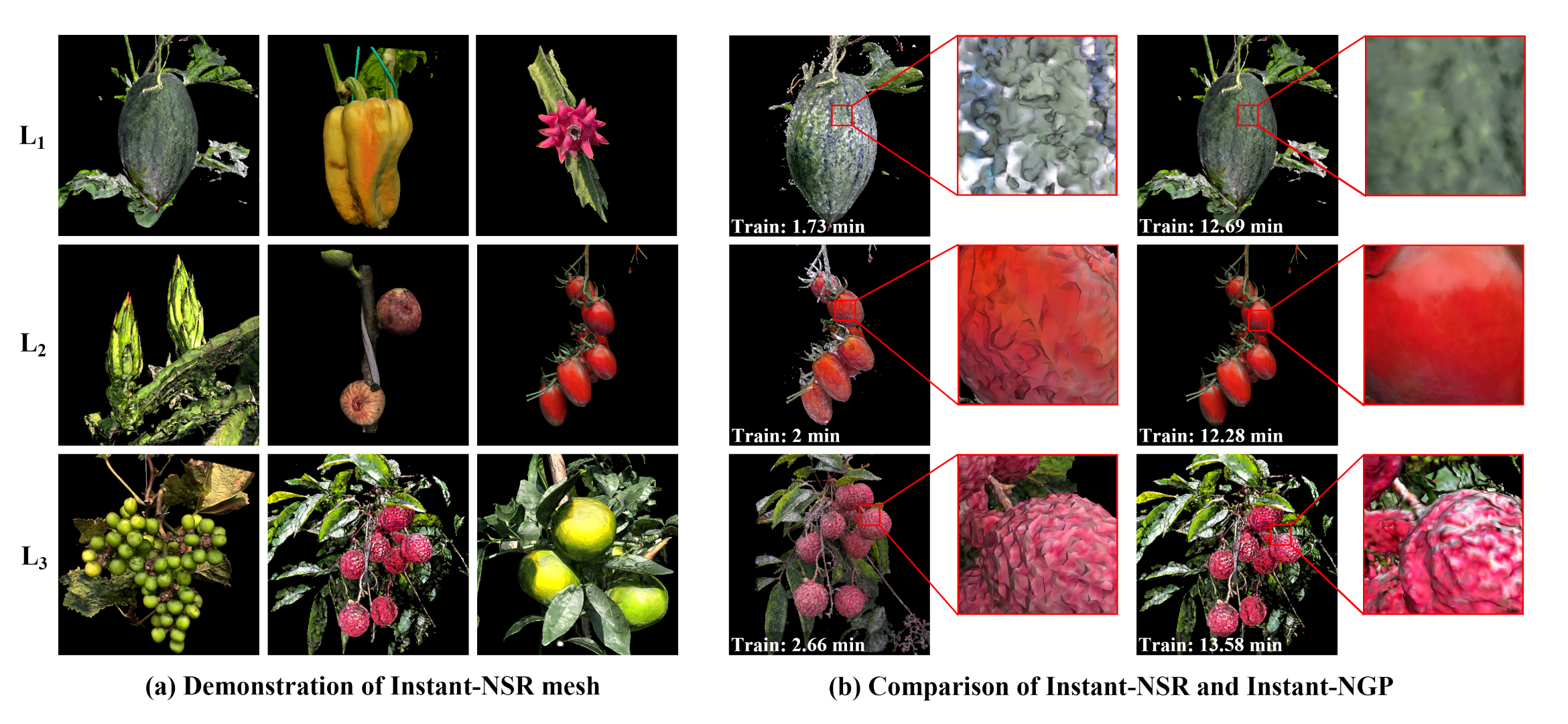}
    \caption{(a) Demonstration of the meshes obtained via Instant-NSR, (b) Comparison of Instant NGP and Instant NSR in details on the model surface. }
    \label{fig:NGP-NSR}
\end{figure*}

Furthermore, Fig. \ref{fig:NGP-NSR} shows the comparison between the mesh of the plants extracted from Instant-NGP and Instant-NSR using the Marching cubes algorithm in Sec. \ref{section:nerf2mesh}. The goal of this controlled experiment is to explore the differences in geometric representation between the NeRF model based on the density architecture and the NeRF model based on the SDF architecture.

\begin{table}[ht]
\centering
\scriptsize
\begin{tabular}{l l|c|c c|c c}
& \textbf{Dataset} & \textbf{RC} & \multicolumn{2}{c|}{\textbf{Instant-NGP}} & \multicolumn{2}{c}{\textbf{Instant-NSR}} \\
& \textbf{Method\textbar{}Metric} & \textbf{Time} & \textbf{Time} & \textbf{PSNR} & \textbf{Time} & \textbf{PSNR} \\
\Xhline{1pt}
 & Watermelon       & 12 min   & 1.73 min & 30.04  & 12.69 min & 29.5   \\
\multirow{-2}{*}{\textbf{L$_1$}}        
                             & Bell Pepper      & 11 min   & 1.71 min & 31.37  & 12.59 min & 24.5   \\
                             & Pitahaya         & 12 min   & 1.80 min & 28.92  & 12.37 min & 29.2   \\
\cline{2-7}
& Pitahaya flower  & 15 min   & 2.16 min & 27.63  & 12.88 min & 26.8   \\
\multirow{-2}{*}{\textbf{L$_2$}}    & Figs     & 14 min   & 1.98 min & 22.46  & 11.39 min & 23.5   \\
                                 & Tomatoes     & 16 min   & 2 min    & 28.25  & 12.28 min & 28.4   \\
\cline{2-7}
& Grapes           & 22 min   & 2.51 min & 29.83  & 13.46 min & 25.5   \\
\multirow{-2}{*}{\textbf{L$_3$}}    & Litchis       & 25 min   & 2.66 min & 27.84  & 13.58 min & 27.5   \\
                             & Oranges          & 21 min   & 2.53 min & 29.89  & 13.47 min & 20.5   \\
\end{tabular}
\caption{Comparison of time and PSNR metrics for different datasets using Reality Capture, Instant-NGP, and Instant-NSR.}
\label{tab:comparison_table}
\end{table}


\subsection{Discussion}
First of all, agricultural scenes, unique in their natural design, present a myriad of challenges when it comes to accurate 3D modeling and analysis. 
\begin{itemize}
    \item \textbf{Natural Diversity and Variances:} Unlike manufactured objects or indoor scenes, nature doesn't adhere to a standardized pattern. This unpredictable variance in terms of plant growth patterns, fruit sizes, leaf orientations, and the positioning of branches makes generalization extremely challenging. 
    \item \textbf{Occlusions and Overlapping: }The growth habit of many plants leads to considerable overlapping and occlusion. Fruits hidden behind leaves, branches intertwining, and dense foliage create visual blockages. These complexities are particularly pronounced in agricultural landscapes, where maximized plant growth is often desired for better yields. 
    \item \textbf{Surface Complexities: }The intricate surface structures, such as the ruggedness of a bark or the delicate vein patterns on leaves, add another layer of complexity. Modeling these minutiae demands high-resolution data capture and advanced algorithms.

\end{itemize}

\subsubsection{Analysis of rendering results}

In terms of efficiency in processing data, one of the key strengths of the Instant-NGP model lies in its efficiency. Across our datasets $L_1$, $L_2$, and $L_3$, the model consistently converges in under three minutes. Furthermore, once converged, the model is capable of real-time rendering from any given viewpoint, demonstrating its prowess in generating novel views. This efficiency exceeds that of traditional RGB camera sampling techniques for phenotype collection, demonstrating the potential of NeRF to enable more versatile and efficient ways of observing plant traits from different perspectives. 

In terms of the quantitative metric PSNR for image rendering, instant-NGP reaches the highest PSNR of 31.37 dB within two minutes for the three scenes in $L_1$. In $L_2$, both datasets (d) and (f) can achieve a PSNR of higher than 26.5 dB. In the scenes of Litchis and grapes in $L_3$, the training time is on average 50 seconds longer than that in $L_1$, but the PSNR also exceeds the baseline of 26.5 dB for all of them. Taking into account the class of the dataset and the quality of the image data, we can see that with simple geometry, instant-NGP can converge very quickly and get a high quality of new view rendering, while complex geometry will cause the network to converge slower. 

\begin{figure}[ht]
    \centering
    \includegraphics[width=1\linewidth]{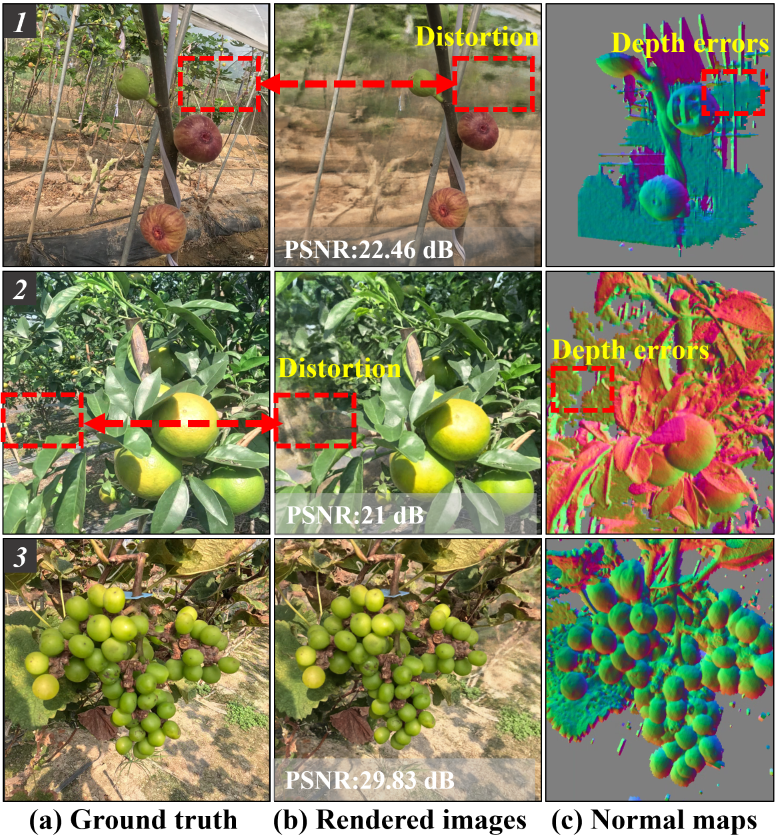}
    \caption{Analysis of quantitative metric PSNR}
    \label{fig:discussion1}
\end{figure}

However, the dataset (e) of figs and (i) of oranges show a very low level of PSNRs: 22.46 dB and 21 dB. Since the PSNR is an average representation of the difference between the rendered image and the ground truth image, we displayed the difference between the full rendered result containing the background and the plant subject and the actual image in Fig.\ref{fig:discussion1}. In this case \textbf{(1) and (2)},  the background in the rendering result is severely defocused, additionally, the background inside differs a lot from the real data, resulting in a low PSNR. On the contrary of \textbf{(3)},the background and the subject are clearly distinguishable and not far from each other, thus giving a high PSNR. From this result, it can be observed that Instant-NGP works well for reconstructing plants in the centre of the scene, but it cannot recover background objects in the distance, which may limit the application of NeRF in large-scale phenotype acquisition.

\subsubsection{Analysis of geometry extraction results}
In the comparison of Instant-NGP and Reality Capture, it is not hard to see that Instant-NGP has a faster modelling speed and at the meantime, it provides a variety of geometrical representations (point cloud, mesh, texture), and even capable of generating rendered images with arbitrary viewpoints. Compared with the traditional Imaging system, NeRF provides a new pattern of information acquisition, which is not to model the plants directly, but to store volumetric and colour information in the neural network, and then convert the parameters of the neural network into the multi-source information needed for phenotyping by methods such as volume rendering. 

However, the NeRF model based on volume density representation lacks geometric constraints to accurately represent the surface of an object, so when modelling plants with smooth surfaces such as watermelons and bell peppers, their meshes are distorted in surface details. The SDF-based NeRF models are excellent at representing the surface of an object with a smooth and continuous iso-surface, but in scenarios where the bump mapping is extremely complex and the SDF gradient varies drastically, the NeRF models converge very slowly.

\section{Conclusion and Future Work} \label{section:conclusion}
In this study, we investigated a novel neural radiance field-based approach for multi-source phenotypic information acquisition to achieve high-fidelity and high-throughput phenotypic reconstruction of a wide range of plants. In our experiments, Instant-NGP is the key method to accelerate the training of NeRF networks, which can model and infer multiple geometric representations faster than the traditional MVS reconstruction method. Moreover, NeRF can generate realistic viewpoint maps through volume rendering. Further, we comprehensively evaluate the performance of NeRF models with two architectures (volume density and SDF) for extracting object surfaces. The experimental results indicate that the volume density-based NeRF model (e.g., Instant-NGP) is suitable for representing plants with uneven surfaces such as litchi, however, the SDF-based one (e.g., Instant-NRS) is more suitable for reconstructing smooth and continuous plant surfaces such as watermelon and grape. 

Future work will focus on enhancing the utility of NeRF by enabling fast and accurate NeRF modelling in the presence of sparse views. Further, we will also investigate how to enhance the background processing while ensuring the reconstruction effect of the subject, which will facilitate the application of NeRF to large-size plant phenotype acquisition.

\bibliographystyle{IEEEtran}
\bibliography{root}

\begin{thebibliography}{10}
\providecommand{\url}[1]{#1}
\csname url@samestyle\endcsname
\providecommand{\newblock}{\relax}
\providecommand{\bibinfo}[2]{#2}
\providecommand{\BIBentrySTDinterwordspacing}{\spaceskip=0pt\relax}
\providecommand{\BIBentryALTinterwordstretchfactor}{4}
\providecommand{\BIBentryALTinterwordspacing}{\spaceskip=\fontdimen2\font plus
\BIBentryALTinterwordstretchfactor\fontdimen3\font minus
  \fontdimen4\font\relax}
\providecommand{\BIBforeignlanguage}[2]{{%
\expandafter\ifx\csname l@#1\endcsname\relax
\typeout{** WARNING: IEEEtran.bst: No hyphenation pattern has been}%
\typeout{** loaded for the language `#1'. Using the pattern for}%
\typeout{** the default language instead.}%
\else
\language=\csname l@#1\endcsname
\fi
#2}}
\providecommand{\BIBdecl}{\relax}
\BIBdecl

\bibitem{sishodia2020applications}
R.~P. Sishodia, R.~L. Ray, and S.~K. Singh, ``Applications of remote sensing in
  precision agriculture: A review,'' \emph{Remote Sensing}, vol.~12, no.~19, p.
  3136, 2020.

\bibitem{fu2020application}
L.~Fu, F.~Gao, J.~Wu, R.~Li, M.~Karkee, and Q.~Zhang, ``Application of consumer
  rgb-d cameras for fruit detection and localization in field: A critical
  review,'' \emph{Computers and Electronics in Agriculture}, vol. 177, p.
  105687, 2020.

\bibitem{feng2021comprehensive}
L.~Feng, S.~Chen, C.~Zhang, Y.~Zhang, and Y.~He, ``A comprehensive review on
  recent applications of unmanned aerial vehicle remote sensing with various
  sensors for high-throughput plant phenotyping,'' \emph{Computers and
  electronics in agriculture}, vol. 182, p. 106033, 2021.

\bibitem{asaari2019analysis}
M.~S.~M. Asaari, S.~Mertens, S.~Dhondt, D.~Inz{\'e}, N.~Wuyts, and
  P.~Scheunders, ``Analysis of hyperspectral images for detection of drought
  stress and recovery in maize plants in a high-throughput phenotyping
  platform,'' \emph{Computers and Electronics in Agriculture}, vol. 162, pp.
  749--758, 2019.

\bibitem{li2020review}
Z.~Li, R.~Guo, M.~Li, Y.~Chen, and G.~Li, ``A review of computer vision
  technologies for plant phenotyping,'' \emph{Computers and Electronics in
  Agriculture}, vol. 176, p. 105672, 2020.

\bibitem{furbank2011phenomics}
R.~T. Furbank and M.~Tester, ``Phenomics--technologies to relieve the
  phenotyping bottleneck,'' \emph{Trends in plant science}, vol.~16, no.~12,
  pp. 635--644, 2011.

\bibitem{rebetzke2019high}
G.~Rebetzke, J.~Jimenez-Berni, R.~Fischer, D.~Deery, and D.~Smith,
  ``High-throughput phenotyping to enhance the use of crop genetic resources,''
  \emph{Plant Science}, vol. 282, pp. 40--48, 2019.

\bibitem{wang2023research}
Y.~Wang, J.~Fan, S.~Yu, S.~Cai, X.~Guo, and C.~Zhao, ``Research advance in
  phenotype detection robots for agriculture and forestry,''
  \emph{International Journal of Agricultural and Biological Engineering},
  vol.~16, no.~1, pp. 14--25, 2023.

\bibitem{zhou2022intelligent}
H.~Zhou, X.~Wang, W.~Au, H.~Kang, and C.~Chen, ``Intelligent robots for fruit
  harvesting: Recent developments and future challenges,'' \emph{Precision
  Agriculture}, vol.~23, no.~5, pp. 1856--1907, 2022.

\bibitem{zhang2018imaging}
Y.~Zhang and N.~Zhang, ``Imaging technologies for plant high-throughput
  phenotyping: a review,'' \emph{Frontiers of Agricultural Science and
  Engineering}, vol.~5, no.~4, pp. 406--419, 2018.

\bibitem{mildenhall2021nerf}
B.~Mildenhall, P.~P. Srinivasan, M.~Tancik, J.~T. Barron, R.~Ramamoorthi, and
  R.~Ng, ``Nerf: Representing scenes as neural radiance fields for view
  synthesis,'' \emph{Communications of the ACM}, vol.~65, no.~1, pp. 99--106,
  2021.

\bibitem{zhao2023phenotyping}
G.~Zhao, R.~Yang, X.~Jing, H.~Zhang, Z.~Wu, X.~Sun, H.~Jiang, R.~Li, X.~Wei,
  S.~Fountas \emph{et~al.}, ``Phenotyping of individual apple tree in modern
  orchard with novel smartphone-based heterogeneous binocular vision and
  yolov5s,'' \emph{Computers and Electronics in Agriculture}, vol. 209, p.
  107814, 2023.

\bibitem{kolhar2023plant}
S.~Kolhar and J.~Jagtap, ``Plant trait estimation and classification studies in
  plant phenotyping using machine vision--a review,'' \emph{Information
  Processing in Agriculture}, vol.~10, no.~1, pp. 114--135, 2023.

\bibitem{kumar2019image}
J.~P. Kumar and S.~Domnic, ``Image based leaf segmentation and counting in
  rosette plants,'' \emph{Information processing in agriculture}, vol.~6,
  no.~2, pp. 233--246, 2019.

\bibitem{ubbens2018use}
J.~Ubbens, M.~Cieslak, P.~Prusinkiewicz, and I.~Stavness, ``The use of plant
  models in deep learning: an application to leaf counting in rosette plants,''
  \emph{Plant methods}, vol.~14, pp. 1--10, 2018.

\bibitem{kang2020fast}
H.~Kang and C.~Chen, ``Fast implementation of real-time fruit detection in
  apple orchards using deep learning,'' \emph{Computers and Electronics in
  Agriculture}, vol. 168, p. 105108, 2020.

\bibitem{jansen2009simultaneous}
M.~Jansen, F.~Gilmer, B.~Biskup, K.~A. Nagel, U.~Rascher, A.~Fischbach,
  S.~Briem, G.~Dreissen, S.~Tittmann, S.~Braun \emph{et~al.}, ``Simultaneous
  phenotyping of leaf growth and chlorophyll fluorescence via growscreen fluoro
  allows detection of stress tolerance in arabidopsis thaliana and other
  rosette plants,'' \emph{Functional Plant Biology}, vol.~36, no.~11, pp.
  902--914, 2009.

\bibitem{clauw2015leaf}
P.~Clauw, F.~Coppens, K.~De~Beuf, S.~Dhondt, T.~Van~Daele, K.~Maleux,
  V.~Storme, L.~Clement, N.~Gonzalez, and D.~Inz{\'e}, ``Leaf responses to mild
  drought stress in natural variants of arabidopsis,'' \emph{Plant physiology},
  vol. 167, no.~3, pp. 800--816, 2015.

\bibitem{dellen2015growth}
B.~Dellen, H.~Scharr, and C.~Torras, ``Growth signatures of rosette plants from
  time-lapse video,'' \emph{IEEE/ACM Transactions on Computational Biology and
  Bioinformatics}, vol.~12, no.~6, pp. 1470--1478, 2015.

\bibitem{humplik2015automated}
J.~F. Humpl{\'\i}k, D.~Laz{\'a}r, A.~Husi{\v{c}}kov{\'a}, and L.~Sp{\'\i}chal,
  ``Automated phenotyping of plant shoots using imaging methods for analysis of
  plant stress responses--a review,'' \emph{Plant methods}, vol.~11, no.~1, pp.
  1--10, 2015.

\bibitem{paulus2019measuring}
S.~Paulus, ``Measuring crops in 3d: using geometry for plant phenotyping,''
  \emph{Plant methods}, vol.~15, no.~1, pp. 1--13, 2019.

\bibitem{kang2022uncertainty}
H.~Kang, Y.~Zang, X.~Wang, and Y.~Chen, ``Uncertainty-driven spiral trajectory
  for robotic peg-in-hole assembly,'' \emph{IEEE Robotics and Automation
  Letters}, vol.~7, no.~3, pp. 6661--6668, 2022.

\bibitem{guo2023improved}
R.~Guo, J.~Xie, J.~Zhu, R.~Cheng, Y.~Zhang, X.~Zhang, X.~Gong, R.~Zhang,
  H.~Wang, and F.~Meng, ``Improved 3d point cloud segmentation for accurate
  phenotypic analysis of cabbage plants using deep learning and clustering
  algorithms,'' \emph{Computers and Electronics in Agriculture}, vol. 211, p.
  108014, 2023.

\bibitem{wu2020mvs}
S.~Wu, W.~Wen, Y.~Wang, J.~Fan, C.~Wang, W.~Gou, and X.~Guo, ``Mvs-pheno: a
  portable and low-cost phenotyping platform for maize shoots using multiview
  stereo 3d reconstruction,'' \emph{Plant Phenomics}, vol. 2020, 2020.

\bibitem{kang2022accurate}
H.~Kang, X.~Wang, and C.~Chen, ``Accurate fruit localisation for robotic
  harvesting using high resolution lidar-camera fusion,'' \emph{arXiv preprint
  arXiv:2205.00404}, 2022.

\bibitem{kang2023semantic}
H.~Kang and X.~Wang, ``Semantic segmentation of fruits on multi-sensor fused
  data in natural orchards,'' \emph{Computers and Electronics in Agriculture},
  vol. 204, p. 107569, 2023.

\bibitem{kok2023obscured}
E.~Kok, X.~Wang, and C.~Chen, ``Obscured tree branches segmentation and 3d
  reconstruction using deep learning and geometrical constraints,''
  \emph{Computers and Electronics in Agriculture}, vol. 210, p. 107884, 2023.

\bibitem{yang20223d}
T.~Yang, J.~Ye, S.~Zhou, A.~Xu, and J.~Yin, ``3d reconstruction method for tree
  seedlings based on point cloud self-registration,'' \emph{Computers and
  Electronics in Agriculture}, vol. 200, p. 107210, 2022.

\bibitem{samavati2023deep}
T.~Samavati and M.~Soryani, ``Deep learning-based 3d reconstruction: A
  survey,'' \emph{Artificial Intelligence Review}, pp. 1--45, 2023.

\bibitem{chen2019learning}
Z.~Chen and H.~Zhang, ``Learning implicit fields for generative shape
  modeling,'' in \emph{Proceedings of the IEEE/CVF Conference on Computer
  Vision and Pattern Recognition}, 2019, pp. 5939--5948.

\bibitem{mescheder2019occupancy}
L.~Mescheder, M.~Oechsle, M.~Niemeyer, S.~Nowozin, and A.~Geiger, ``Occupancy
  networks: Learning 3d reconstruction in function space,'' in
  \emph{Proceedings of the IEEE/CVF conference on computer vision and pattern
  recognition}, 2019, pp. 4460--4470.

\bibitem{park2019deepsdf}
J.~J. Park, P.~Florence, J.~Straub, R.~Newcombe, and S.~Lovegrove, ``Deepsdf:
  Learning continuous signed distance functions for shape representation,'' in
  \emph{Proceedings of the IEEE/CVF conference on computer vision and pattern
  recognition}, 2019, pp. 165--174.

\bibitem{bp1984ray}
K.~J. V.~H. BP, ``Ray tracing volume densities acm siggraph comput,''
  \emph{Graph}, vol.~18, no.~3, p. 165, 1984.

\bibitem{lorensen1998marching}
W.~E. Lorensen and H.~E. Cline, ``Marching cubes: A high resolution 3d surface
  construction algorithm,'' in \emph{Seminal graphics: pioneering efforts that
  shaped the field}, 1998, pp. 347--353.

\bibitem{sander2001texture}
P.~V. Sander, J.~Snyder, S.~J. Gortler, and H.~Hoppe, ``Texture mapping
  progressive meshes,'' in \emph{Proceedings of the 28th annual conference on
  Computer graphics and interactive techniques}, 2001, pp. 409--416.

\bibitem{levy2023least}
B.~L{\'e}vy, S.~Petitjean, N.~Ray, and J.~Maillot, ``Least squares conformal
  maps for automatic texture atlas generation,'' in \emph{Seminal Graphics
  Papers: Pushing the Boundaries, Volume 2}, 2023, pp. 193--202.

\bibitem{muller2022instant}
T.~M{\"u}ller, A.~Evans, C.~Schied, and A.~Keller, ``Instant neural graphics
  primitives with a multiresolution hash encoding,'' \emph{ACM Transactions on
  Graphics (ToG)}, vol.~41, no.~4, pp. 1--15, 2022.

\bibitem{wang2021neus}
P.~Wang, L.~Liu, Y.~Liu, C.~Theobalt, T.~Komura, and W.~Wang, ``Neus: Learning
  neural implicit surfaces by volume rendering for multi-view reconstruction,''
  \emph{arXiv preprint arXiv:2106.10689}, 2021.

\bibitem{zhao2022human}
F.~Zhao, Y.~Jiang, K.~Yao, J.~Zhang, L.~Wang, H.~Dai, Y.~Zhong, Y.~Zhang,
  M.~Wu, L.~Xu \emph{et~al.}, ``Human performance modeling and rendering via
  neural animated mesh,'' \emph{ACM Transactions on Graphics (TOG)}, vol.~41,
  no.~6, pp. 1--17, 2022.

\bibitem{schonberger2016structure}
J.~L. Schonberger and J.-M. Frahm, ``Structure-from-motion revisited,'' in
  \emph{Proceedings of the IEEE conference on computer vision and pattern
  recognition}, 2016, pp. 4104--4113.

\bibitem{mildenhall2019local}
B.~Mildenhall, P.~P. Srinivasan, R.~Ortiz-Cayon, N.~K. Kalantari,
  R.~Ramamoorthi, R.~Ng, and A.~Kar, ``Local light field fusion: Practical view
  synthesis with prescriptive sampling guidelines,'' \emph{ACM Transactions on
  Graphics (TOG)}, vol.~38, no.~4, pp. 1--14, 2019.

\end{thebibliography}
\end{document}